\documentclass[a4paper,fleqn,authoryear]{cas-sc}
\usepackage[authoryear]{natbib}
\usepackage{graphicx}
\usepackage{adjustbox}
\usepackage{booktabs}
\usepackage{multirow}
\usepackage{makecell}
\usepackage{xcolor}
\usepackage{colortbl}
\usepackage{array}
\usepackage{amsmath,amssymb,amsfonts}
\usepackage{algorithm}
\usepackage{algorithmic}
\usepackage{soul}
\usepackage{pifont}
\usepackage{pgfplots}
\pgfplotsset{compat=1.18}
\usepackage{textcomp}
\usepackage{arydshln}
\usepackage{url}
\usepackage{hyperref}
\usepackage{float}
\floatstyle{plain}
\restylefloat{figure}
\floatstyle{plaintop}
\restylefloat{table}
\begin{document}
\let\WriteBookmarks\relax
\renewcommand{\floatpagefraction}{0.7}
\renewcommand{\textfraction}{0.1}
\renewcommand{\topfraction}{0.9}
\renewcommand{\bottomfraction}{0.8}
\setcounter{topnumber}{3}
\setcounter{totalnumber}{5}

\shorttitle{Agri-CPJ: A Training-Free Explainable Framework for Agricultural Pest Diagnosis}

\shortauthors{Anonymous}

\title[mode=title]{Agri-CPJ: A Training-Free Explainable Framework for Agricultural Pest Diagnosis Using Caption-Prompt-Judge and LLM-as-a-Judge}




\author[1]{Wentao Zhang}
\author[2]{Qi Zhang}
\author[3]{Mingkun Xu}
\author[4]{Mu You}
\author[4]{Henghua Shen}
\author[4]{Zhongzhi He}
\author[4]{Keyan Jin}
\author[5]{Derek F. Wong}
\author[4]{Tao Fang\corref{cor1}}
\affiliation[1]{organization={Business School, Shandong University of Technology},
                city={Shandong},
                country={China}}

\affiliation[2]{organization={Faculty of Data Science, City University of Macau},
                city={Macau SAR},
                country={China}}

\affiliation[3]{organization={Guangdong Institute of Intelligent Science and Technology},
                city={Zhuhai},
                country={China}}

\affiliation[4]{organization={Macau Millennium College},
                city={Macau SAR},
                country={China}}
                
\affiliation[5]{organization={Department of Computer and Information Science, University of Macau},
                city={Macau SAR},
                country={China}}

\cortext[cor1]{Corresponding Author: Tao Fang (taofang@mmc.edu.mo)}

\begin{abstract}
Crop disease diagnosis from field photographs faces two recurring problems: models that score well on benchmarks frequently hallucinate species names, and when predictions are correct, the reasoning behind them is typically inaccessible to the practitioner. This paper describes Agri-CPJ (Caption-Prompt-Judge), a training-free few-shot framework in which a large vision-language model first generates a structured morphological caption, iteratively refined through multi-dimensional quality gating, before any diagnostic question is answered. Two candidate responses are then generated from complementary viewpoints, and an LLM judge selects the stronger one based on domain-specific criteria. Caption refinement is the component with the largest individual impact: ablations confirm that skipping it consistently degrades downstream accuracy across both models tested. On CDDMBench, pairing GPT-5-Nano with GPT-5-mini-generated captions yields \textbf{+22.7} pp in disease classification and \textbf{+19.5} points in QA score over no-caption baselines. Evaluated without modification on AgMMU-MCQs, GPT-5-Nano reached 77.84\% and Qwen-VL-Chat reached 64.54\%, placing them at or above most open-source models of comparable scale despite the format shift from open-ended to multiple-choice. The structured caption and judge rationale together constitute a readable audit trail: a practitioner who disagrees with a diagnosis can identify the specific caption observation that was incorrect. Code and data are publicly available.\footnote{\url{https://github.com/CPJ-Agricultural/CPJ-Agricultural-Diagnosis}}
\end{abstract}


\maketitle

\section{Introduction}
\label{sec:introduction}

Accurate crop disease diagnosis from field imagery requires two properties that most current systems do not simultaneously provide: correct identification of the pathogen or condition, and an explanation of that identification in terms a practitioner can evaluate. Crop disease spreads rapidly, and an incorrect or unverifiable recommendation---an erroneous fungicide prescription, or no actionable guidance at all---carries direct economic consequences. These twin requirements motivate the framework described here.

Current mainstream solutions lean heavily on unimodal visual pipelines such as CNN classifiers~\citep{duro2012comparison} and YOLO-style detectors~\citep{badgujar2024agricultural}. These models work well on the benchmarks they were trained for. In our experience, however, they break down when moved even slightly out of distribution---different lighting, a new crop variety, a slightly unusual growth stage---and when they do fail, they offer no clue as to why. What such a system returns is, at best, a class label. For an agronomist weighing fungicide application costs against expected crop loss, or an extension officer advising smallholders across several districts, a single predicted category provides no basis for verifying the conclusion or adjusting the recommendation to local conditions.

\begin{figure}[!t]
    \centering
    \includegraphics[width=0.95\textwidth]{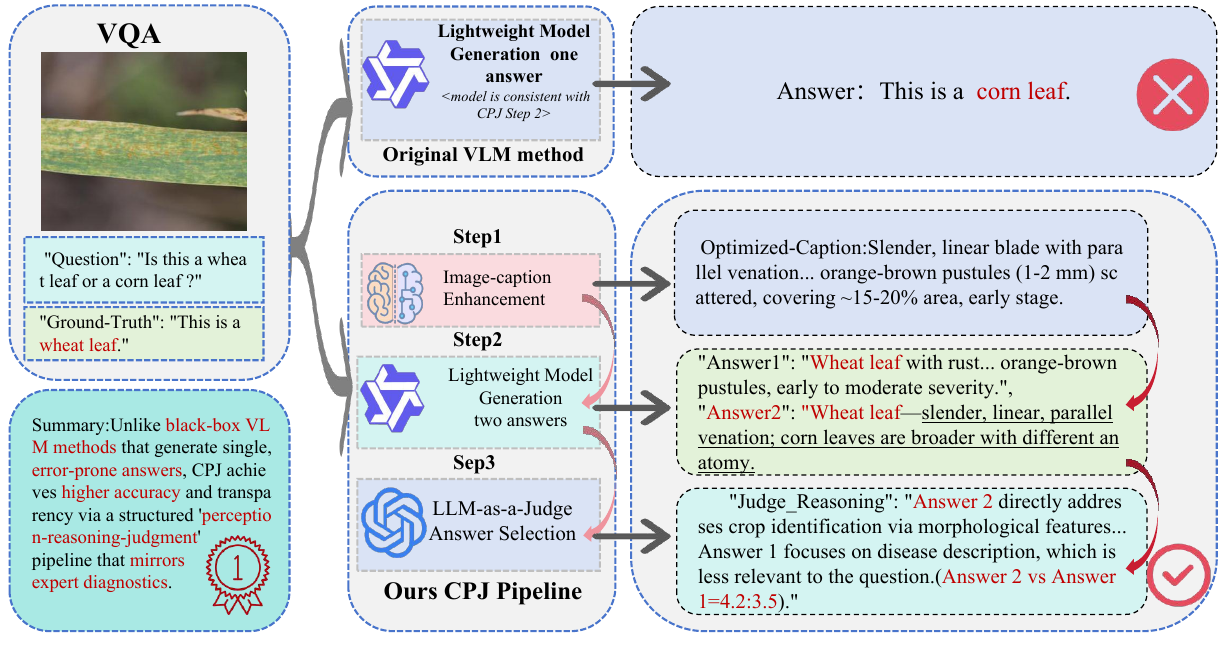}
    \caption{Comparison between traditional VLM and Agri-CPJ framework on crop species identification. Traditional VLMs provide direct but error-prone black-box predictions (incorrectly identifying wheat as corn, \ding{55}), lacking explicit reasoning chains. In contrast, Agri-CPJ delivers accurate diagnoses (\ding{51}) through a structured perception-reasoning-judgment pipeline: (i) explicit feature extraction via optimized captions (e.g., ``slender linear blade with parallel venation''), (ii) dual complementary answer generation addressing morphological characteristics, and (iii) LLM-as-a-Judge selection based on diagnostic validity. This expert-mimicking approach significantly enhances both accuracy and interpretability.}
    \label{fig:comparison}
\end{figure}

Large Vision-Language Models---Qwen-VL~\citep{bai2023qwen}, the GPT series~\citep{achiam2023gpt}, LLaVA~\citep{liu2023visual}, Flamingo~\citep{alayrac2022flamingo}---represent a qualitative shift in capability. Unlike CNN classifiers, they accept free-form diagnostic questions and, under favourable conditions, articulate some reasoning alongside an answer. Agricultural VQA studies~\citep{zhang2024visual,li2023blip} confirm that they produce richer outputs than label-only pipelines. The difficulty is consistency: without any mechanism compelling the model to describe observable features before reaching a conclusion, the same image can yield correct or incorrect answers depending on subtle prompt variation. Fig.~\ref{fig:comparison} (top) shows a representative failure---a wheat leaf identified as corn, with no intermediate output that would let a user locate where the reasoning broke down.

Fine-tuning on labelled agricultural data is the standard response to this, and several groups have pursued it~\citep{cao2023cucumber,radford2021learning,alayrac2022flamingo}. Three difficulties recur in practice. Annotated crop disease datasets are costly to assemble and tend to reflect specific geographic and varietal conditions; \citet{saadati2024out} document sharp accuracy drops when models encounter out-of-distribution species. Interpretability does not improve either---gradient attribution maps reveal which image regions influenced a prediction, but not the observational logic behind it.

Our goal was therefore a system that generalises across crop types and disease categories without retraining, and that exposes its reasoning in a form that domain experts can scrutinise and, where necessary, dispute at the level of individual observations.

There is a second, underappreciated problem with fine-tuning in agricultural deployment contexts. The model's knowledge is frozen at training time. When a previously unrecorded pathogen variant is first confirmed, or when a grower presents a crop variety whose symptom presentation differs from anything in the training set, a fine-tuned system requires retraining before it can respond reliably---a process demanding labelled data, compute infrastructure, and engineering time that most cooperative extension services cannot readily mobilise. Prompting-based approaches handle novel cases differently: a new crop type or disease can be incorporated as a few-shot example on the day it is identified, without modifying any model parameter. In a slow-moving situation, that distinction is a convenience; during an outbreak spreading faster than a retraining cycle can complete, it determines whether the diagnostic tool is deployable at all.

Work on logic-driven anomaly detection~\citep{jin2025logicad,ye2025vera,song2026instance,xu2025towards} suggests an alternative design principle: rather than optimising a model for a fixed task distribution, externalise the reasoning so that each inference step can be independently inspected and corrected. \textbf{Agri-CPJ}
\footnote{This manuscript substantially extends \textcolor{blue}{Our Prior ICASSP 2026 paper~\cite{zhang2025cpj}} from 4 to 20+ pages, presenting a well-structured and principled framework, extensive experiments, and deeper insights.}
applies this principle to crop disease. Between a leaf photograph and a diagnostic conclusion lies a gap. In most current systems that gap is crossed by a single forward pass with no accessible intermediate state. We require instead that it be traversed through a text record of morphological observations---one a practitioner can read, question, or correct before accepting any downstream answer.


\begin{figure}[!t]
    \centering
    \includegraphics[width=0.95\textwidth]{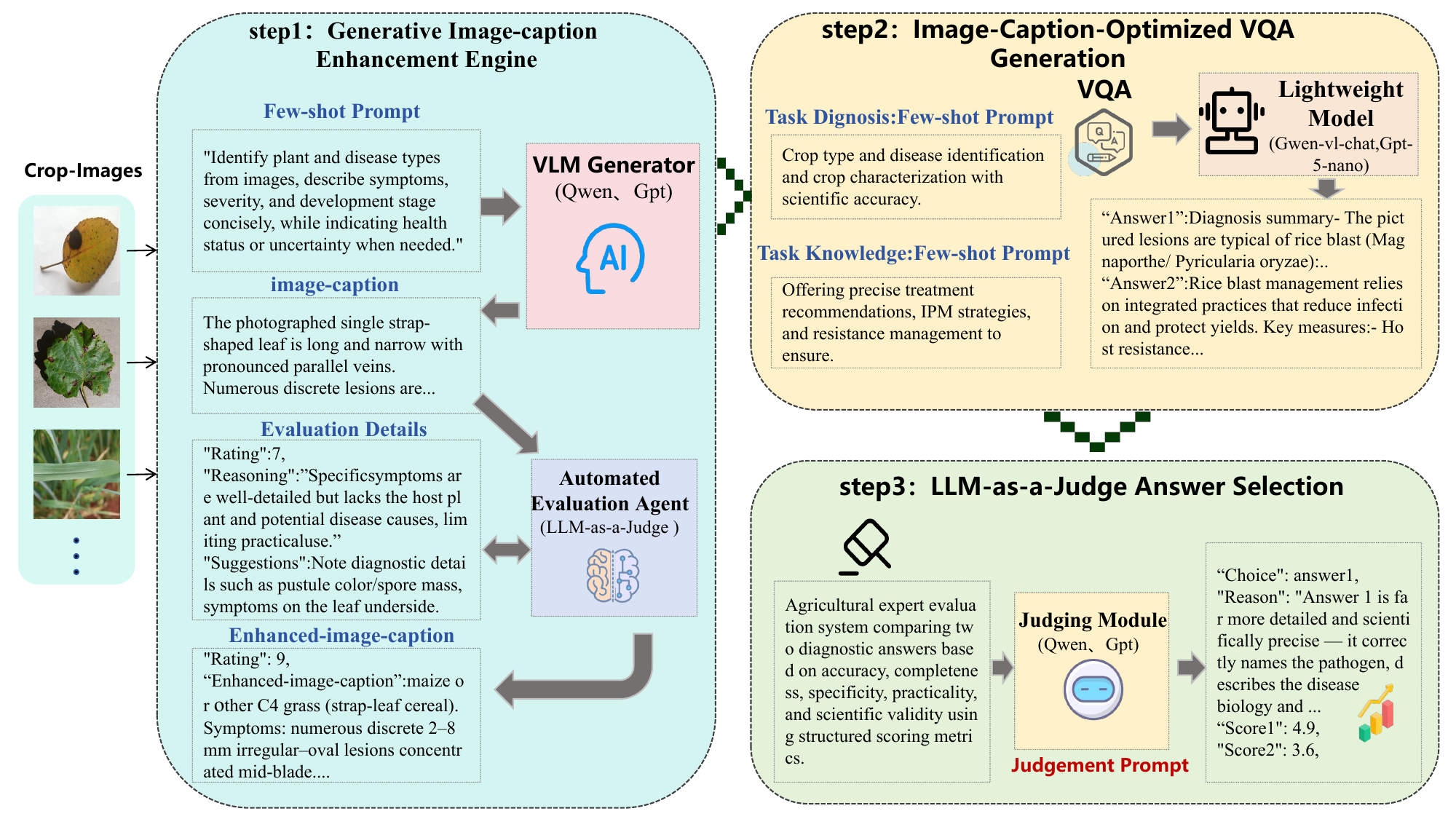}
    \caption{Overview of the ``Agri-CPJ'' pipeline for explainable Agri-Pest VQA, featuring three cohesive stages: (i) \textbf{Generative Explanational Captioning}: A large vision-language model analyzes input crop disease images to generate multi-angle descriptive captions detailing plant morphology, disease symptoms, severity levels, and diagnostic uncertainties. Each generated caption is then passed through an iterative LLM-as-a-Judge refinement loop until it satisfies quality thresholds for observational accuracy, completeness, and diagnostic neutrality. (ii) \textbf{Task-Specific Prompt-Based VQA Generation}: The refined caption is combined with the original image and few-shot exemplars to construct the VQA prompt; the model generates two candidate responses from distinct diagnostic perspectives---one targeting disease recognition (crop identity, symptom characterisation, visible indicators) and one targeting management guidance (treatment protocols, prevention measures). (iii) \textbf{LLM-as-a-Judge Answer Selection}: A more capable LLM scores both candidates against a reference answer using multiple criteria (factual correctness, completeness, specificity, practical applicability), selects the stronger response, and returns an explicit rationale documenting the basis for the selection.}
    \label{fig:framework}
\end{figure}

The framework we propose, \textbf{Agri-CPJ}, has three stages. A large LVLM opens by reading the input image and writing a structured caption---covering morphology, symptoms, severity, and uncertainty---while deliberately withholding crop and disease names to avoid encoding a premature label. That caption is critiqued by an LLM judge and rewritten if it falls short of a quality threshold. The refined caption then travels alongside the original image to a lightweight VQA model, which produces two candidate answers from different diagnostic angles: one disease-focused, one crop-focused. A second judge LLM scores both against a reference and picks the stronger one, documenting why. Fig.~\ref{fig:comparison} (bottom) shows the result: the wheat/corn confusion is corrected because the right morphological evidence was on the table before any label was attached.

CPJ adjusts to new cases through the context window: a newly confirmed disease variant or crop type enters the few-shot exemplars on the same day it is identified, without modifying any model parameter. The ceiling, however, is set by the vision encoder's perceptual resolution. For photographs taken during the earliest infection stages---when surface symptoms are minimal or ambiguous---even a carefully constructed caption cannot supply discriminative information the image does not contain.

The interpretability claim merits careful statement. Gradient-based attribution maps and attention visualisations indicate which image regions a model weighted---they do not describe what the model understood those regions to mean, and most field practitioners lack the background to translate pixel-level saliency scores into agronomic reasoning. CPJ operates on different ground. The structured caption is a text document: a plant pathologist can read it, assess whether the morphological observations are accurate, and identify specific errors before examining any downstream answer. Both candidate responses and the judge's selection rationale are similarly readable. A practitioner who disagrees with the system's conclusion can point to the particular caption observation that seems wrong, argue that one candidate better fits the field situation, or note that the judge's preference conflicts with local experience---and do all of this without understanding how the underlying model weights are configured. Saliency maps offer no equivalent affordance. A practitioner cannot point to a specific pixel cluster that caused a wrong prediction and explain, in observational terms, why it should have been read differently. In Agri-CPJ, the caption is that target: disagreement has a named location and a readable description.

The numbers support these claims across both benchmarks. On CDDMBench~\citep{liu2024multimodal}, the GPT-5-Nano backbone paired with GPT-5-mini-generated captions yields \textbf{+22.7} pp in disease classification accuracy and \textbf{+19.5} points on the knowledge QA sub-task relative to no-caption zero-shot baselines. Evaluated without modification on AgMMU-MCQs~\citep{gauba2025agmmu}---a structurally distinct multiple-choice format spanning five agricultural task categories---the full CPJ configuration (Caption + Few-shot + LLM-as-a-Judge) achieves 77.84\% with GPT-5-Nano and 64.54\% with Qwen-VL-Chat, placing it at or above most open-source models of comparable parameter scale.

This paper makes three contributions. First, we present Agri-CPJ, a training-free few-shot pipeline in which visual reasoning is externalised as an iteratively refined caption before any diagnostic question is answered, enabling practitioners to trace a conclusion back to the specific image observations that supported it. Second, the VQA stage generates two candidate responses from complementary diagnostic viewpoints, with caption prompts withholding crop and disease names to suppress premature label commitment; a multi-criteria LLM judge selects between candidates and produces an explicit rationale. Third, we report controlled experiments on CDDMBench and AgMMU-MCQs across two backbone architectures, together with a human validation study in which LLM judge selections agreed with expert plant pathologist assessments at $\kappa = 0.88$.

\section{Related Work}
\label{sec:related}

\subsection{Agricultural Vision-Language Models}

Agricultural image analysis has followed the broad trajectory of computer vision: hand-crafted descriptors~\citep{kumar2015plant} gave way to convolutional architectures, which have since been largely superseded by Vision Transformers~\citep{dosovitskiy2020image} and language-aligned backbones~\citep{yadav2023comparative}; \citet{zhu2024harnessinglargevisionlanguage} survey this transition in detail. Within-distribution accuracy has improved substantially---purpose-built models such as AgriGPT-VL~\citep{yang2025agrigptvlagriculturalvisionlanguageunderstanding} and AgriVLM~\citep{yu2024framework} report figures above 90\% on their own test splits, achieved through parameter-efficient fine-tuning. What benchmark tables tend to leave unreported is cross-distribution behaviour: whether the same model holds up when applied to a crop variety absent from training, or to field photographs taken under lighting and soil conditions that differ from the dataset.

Sensor fusion attempts to address the contextual shallowness of image-only methods: Rice-Fusion~\citep{patil2022rice} pairs leaf images with agro-meteorological readings, and \citet{lu2024application} incorporate image, text, and sensor streams together. For knowledge breadth, Agri-LLaVA~\citep{wang2024agri} covers 221 pest and disease categories via 400K training entries. The closest precedent to our approach is WDLM~\citep{zhang2024visual}, which generates step-by-step explanations for wheat disease rather than terminal labels---though it still requires disease-specific fine-tuning and does not support iterative refinement of those explanations.

The benchmark picture is less flattering than the model papers suggest. AgroBench~\citep{Shinoda_2025_ICCV} finds that more than half of all VLM errors come from missing domain knowledge, not poor vision. AgEval~\citep{arshad2025leveraging} and VL-PAW~\citep{yu2025vl} show modest few-shot and fine-tuning gains, but hallucination persists~\citep{ren2025performance} and truly lightweight deployable models are still rare~\citep{vu2025agridetectvl}. The availability of large-scale datasets---MIRAGE~\citep{dongre2025mirage}, CDDMBench~\citep{liu2024multimodal}, AgMMU~\citep{gauba2025agmmu}---has helped in-domain results~\citep{yang2024application,ranario2025vision,su2022survey}, but models trained on them tend to overfit narrow distributions~\citep{saadati2024out} and rarely say \textit{why} they reached a conclusion.

\subsection{Explainable AI in Vision-Language Models}

The interpretability problem in VLMs is older than VLMs themselves. Gradient-based attribution methods---Shapley values~\citep{dinga2025you}, LRP/LIME~\citep{kamal2025explainable}, GEST~\citep{masala2025vision}, VALE~\citep{natarajan2024vale}---were the default for years. The objection to them, which we share, is not that they are technically wrong but that a heat map over pixel regions is not the same thing as a justification a domain expert can follow. A plant pathologist does not think in gradients; they describe what they observe and reason from those descriptions.

Chain-of-Thought methods move in the right direction. Multimodal-CoT~\citep{zhang2023multimodal} separates rationale generation from answer inference, which already reduces hallucination measurably. DDCoT~\citep{zheng2023ddcot} goes further by splitting the reasoning work between language and vision submodules. ICoT~\citep{gao2025interleaved} interleaves visual and textual tokens step by step, achieving impressive results but at considerable architectural cost. CPJ takes a structurally more direct route: the visual observation is externalised as a standalone text document before any question-answering step is initiated. Because this caption exists as a discrete, human-readable artifact, it can be reviewed, corrected, or substituted independently of the underlying model---without any architectural modification.

Among retrieval-augmented methods, REVEAL~\citep{hu2023reveal} is the closest in motivation: it grounds model outputs in a curated external knowledge base so that responses can be traced to specific sources. The operational gap relative to Agri-CPJ is that REVEAL depends on constructing and maintaining that knowledge base prior to deployment, whereas our captions are produced directly from the input image at inference time and require no external corpus. CoT-VLA~\citep{zhao2025cot} applies a comparable reasoning-before-action principle to robotic manipulation, providing evidence that externalised intermediate representations confer benefits across task domains beyond visual classification.

\subsection{LLM-as-a-Judge for Quality Assurance}

LLM-based evaluation has accumulated a well-documented set of failure modes. MT-Bench~\citep{zheng2023judging} provided an early systematic characterisation: overall human--LLM agreement exceeded 80\%, but two biases emerged consistently---position bias, whereby judges assign higher scores to whichever candidate appears first in the prompt, and verbosity bias, whereby longer responses receive preferential treatment regardless of content. JudgeLM~\citep{zhu2023judgelm} reduces both through swap augmentation during judge fine-tuning. PolyVis~\citep{padarha2025evaluating} introduces a more fundamental concern: across multilingual and cross-domain settings, judge reliability degrades in ways that model scale alone does not explain, indicating that evaluation quality is contingent on domain alignment rather than general capability.

For multimodal settings, reliability is less settled. \citet{chen2024mllm} show LLM judges are more consistent at pairwise comparison than absolute scoring; Agri-CPJ uses pairwise comparison partly for this reason. Prometheus-Vision~\citep{lee2024prometheus} introduced open-source multimodal judging with configurable rubrics, and Woodpecker~\citep{yin2024woodpecker} demonstrated that hallucinations can be corrected through staged validation without any additional training. \citet{shi2025judging} measured scoring gaps of up to five points from human raters, with verbosity bias severe enough to change the selected answer. We encountered this in early Agri-CPJ runs: the judge consistently preferred longer answers even when shorter ones were more factually accurate. Removing response length from the scoring criteria brought verbosity-driven misselections from roughly 12\% down to 5\%.

\section{Methodology}
\label{sec:methodology}

Rather than building an end-to-end model, Agri-CPJ decomposes diagnosis into three stages that a human expert might recognize: look carefully at the image and describe what you see; reason over those observations to form candidate answers; then critically compare the candidates and pick the stronger one. Each stage is implemented by prompting off-the-shelf LVLMs, so no gradient updates are ever required.

\subsection{Framework Design Philosophy}
\label{subsec:design_philosophy}

Examining incorrect diagnoses produced by baseline LVLMs, we found a recurring pattern: the model had frequently encoded relevant visual information---lesion shape, pustule colour, venation geometry---but had not committed those observations to any retrievable representation before generating an answer. The diagnostic error was therefore not a failure of visual processing per se, but a failure to constrain the reasoning with explicit observational evidence. This finding motivated the central architectural decision in Agri-CPJ: the visual observation must be externalised as a text artifact, accessible to both the downstream model and any human reviewer, before any diagnostic question is posed.

A separate problem emerged once the captioning stage was in place. A single VQA response tends to concentrate on whichever diagnostic dimension the model finds most tractable---disease identification at the expense of crop morphology, or the reverse. Generating two responses from distinct viewpoints before any selection step addressed this: even without the judge, the pair jointly covered substantially more of the relevant diagnostic information than any individual response. The computational overhead is limited to one additional VQA call per query, and the improvement in information coverage was consistent across development experiments.

The third requirement was auditability at the point of decision. When a practitioner disagrees with a system output, the disagreement should be articulable---directed at a specific observation in the caption, a specific claim in one candidate answer, or a specific aspect of the judge's scoring rationale---rather than expressed only as a general distrust of the result. Every Agri-CPJ output therefore includes the refined caption with its quality score, both candidate answers with individual scores, and the judge's written justification for its selection. Each of these components constitutes a locatable point of intervention. The architecture is defined by these three structural commitments: externalised captions, dual-perspective answer generation, and scored selection accompanied by an explicit rationale.

The distinction from end-to-end approaches is clearest by contrast. A fine-tuned model achieving strong benchmark accuracy has merged observation, inference, and output into a single forward pass; when it errs, no intermediate representation survives that a reviewer could examine. Chain-of-thought prompting moves closer to what Agri-CPJ does---it generates a reasoning trace before the final answer---but the trace and the answer are produced together in one decoding sequence, which allows the reasoning to be constructed retrospectively around a conclusion already implicit in the generation. Agri-CPJ separates these steps in a harder sense: the caption is generated and quality-checked before the diagnostic question is even presented to the VQA model, so the observation stage is genuinely prior to and independent of the reasoning stage. An error in the caption does not propagate silently; it is visible in the text before any answer is produced.

One operational consequence of this design is caption reusability. Since the caption is produced before any question is answered, a single refined caption can serve multiple subsequent queries about the same image. In a multi-turn diagnostic session---where a grower uploads a photograph and asks in sequence about disease identity, harvest safety, and appropriate fungicide---the iterative refinement step runs once and the resulting caption is passed to the VQA and judge stages for each question independently. Given that caption refinement accounts for the majority of API cost per session (it requires the vision-capable model and potentially multiple iterations), amortising it across several queries substantially reduces the effective per-question cost in real deployment scenarios.

\subsection{Image Interpretability and Caption Enhancement}
\label{subsec:caption_enhancement}

Agricultural pathology relies on a specific diagnostic vocabulary---chlorotic halo geometry, lesion boundary regularity, pustule colour and distribution---that sits at some distance from the feature representations computed by general-purpose vision encoders. When a model proceeds directly from pixel input to a disease label, this vocabulary gap is bridged implicitly inside a single forward pass, and any error in that translation has no externally visible form. An explicit caption interposes a translatable record: the model commits its visual interpretation to text before any diagnostic step, and that text can be checked against field observation, corrected if inaccurate, or replaced with an expert-authored description without touching the downstream architecture.

In our validation, agronomists who ultimately disagreed with a final diagnosis frequently confirmed that the caption's morphological description was accurate. This indicated that the error resided in the interpretive step---how the model moved from observed morphology to a pathogen class---rather than in the visual encoding itself. That is a more tractable failure mode: correcting a misinterpretation in a text document is substantially more accessible than diagnosing and adjusting a hidden-layer representation.

One prompt engineering decision required substantially more iteration than anticipated: captions are constrained to exclude crop species names and disease labels entirely. The rationale is that even indirect diagnostic vocabulary---phrases such as ``distribution consistent with a biotrophic pathogen'' or ``lesion morphology typical of late blight''---effectively anchors the downstream VQA model to a pathogen class before it has examined the full observational evidence, introducing a form of label leakage through the caption channel. Arriving at prompt formulations that reliably elicited morphology-focused, nomenclature-free descriptions took approximately twelve revision rounds across development. Even with the final prompts, roughly 35\% of first-pass captions in our validation set carried at least one implicit diagnostic cue; the quality-gating refinement loop was specifically designed to detect and remove these.

Caption generation is formalised in Eq.~\eqref{eq:caption}: a large vision-language model receiving image $I$ alongside few-shot prompt $P_{\text{few}}$ produces an initial description $C_0$. The exemplars in $P_{\text{few}}$ are chosen to demonstrate the target caption register---precise morphological observation, symptom coverage, severity estimation---while scrupulously avoiding any crop species name or disease label, so that the debiasing constraint is enforced through demonstration rather than instruction alone.
\begin{equation}
C_0 = \mathcal{M}_{\text{VLM}}(I, P_{\text{few}}),
\label{eq:caption}
\end{equation}

Quality control happens through multi-dimensional scoring across $k$ dimensions $\mathcal{D} = \{d_1, \ldots, d_k\}$ covering accuracy, completeness, specificity, relevance, and clarity:
\begin{equation}
s(C) = \frac{1}{k} \sum_{i=1}^{k} w_i \cdot d_i(C),
\label{eq:quality_score}
\end{equation}
Captions that fall below a threshold $\tau$ go back to the VLM with targeted feedback $\mathcal{R}(\cdot)$ on the specific deficiencies:
\begin{equation}
C^\ast =
\begin{cases}
C_0, & \text{if } s(C_0) \geq \tau, \\
\mathcal{M}_{\text{VLM}}(I, \mathcal{R}(C_0)), & \text{otherwise.}
\end{cases}
\label{eq:refinement}
\end{equation}
We set $\tau = 8.0/10.0$ across all experiments. Raising it to 8.5 tightened quality slightly but roughly doubled the average number of refinement iterations and the associated API cost, so we kept the lower threshold. In the rare cases where the loop does not converge within $N_{\max}$ iterations, we use the best caption produced so far rather than failing the query entirely. Complete system prompts and few-shot exemplars for all three stages are provided in Appendix~\ref{appendix:prompts}.

\subsection{Explanational Caption-Optimized VQA}
\label{subsec:vqa}

A single VQA pass over the refined caption tends to anchor on whichever diagnostic dimension the model finds easiest---often the disease label, at the expense of crop morphology or management guidance. Generating two candidate answers from explicitly different viewpoints is a simple way to guard against this. Formally, for viewpoints $\mathcal{V}_1$ and $\mathcal{V}_2$:
\begin{equation}
A^{(i)} = \mathcal{M}_{\text{VQA}}(I, C^\ast, Q \mid \mathcal{V}_i), \quad i \in \{1, 2\},
\label{eq:dual_answer}
\end{equation}
so that the candidate set becomes $\mathcal{A} = \{A^{(1)}, A^{(2)}\} = \mathcal{M}_{\text{VQA}}(X, P_{\text{task}})$, where $X = (I, C^\ast, Q)$.
\begin{equation}
\mathcal{A} = \{A^{(1)}, A^{(2)}\} = \mathcal{M}_{\text{VQA}}(X, P_{\text{task}}),
\label{eq:vqa}
\end{equation}
The specific viewpoint split we settled on is disease-focused vs.\ crop-focused for recognition questions, and treatment-protocol vs.\ disease-mechanism for knowledge questions. We tried several other pairings in preliminary experiments---e.g.\ severity-focused vs.\ management-focused---but found that the disease/crop split gave the most complementary outputs on CDDMBench; the two answers rarely overlapped heavily, which gave the judge a meaningful choice.

Agricultural diagnosis is structurally multi-dimensional. A query about a diseased maize leaf typically requires both a pathogen identification and an assessment of the crop's developmental stage and associated susceptibility---two perspectives that draw on distinct portions of the caption content. The disease-focused candidate reasons from symptom patterns toward pathogen class and treatment protocol; the crop-focused candidate reasons from morphological identity toward growth stage and vulnerability profile. Generating these two paths separately, rather than producing a single response that conflates them, ensures that the judge evaluates candidates with genuinely different information content rather than candidates that differ mainly in surface phrasing.

The viewpoint pairing differs by task category. For recognition tasks---determining what organism or condition is present---the two candidates address: (i) disease or pest identification, emphasising lesion morphology, symptom severity, and characteristic distribution patterns; and (ii) crop identification, emphasising plant architecture, variety-specific features, and structural markers. For knowledge tasks---explaining mechanisms or recommending management---the pairing shifts to: (i) treatment, prevention, and integrated control strategies with specified methods and timing; and (ii) disease aetiology, pathogen lifecycle, and environmental conditions governing spread. This split maintains architectural consistency across task types while directing each perspective toward the information most relevant to that category of question.

Few-shot prompting populates $P_{\text{task}}$ with in-context exemplars $\mathcal{E} = \{(I_j, C_j, Q_j, A_j)\}_{j=1}^{n}$:
\begin{equation}
P(A \mid I, C, Q, \mathcal{E}) \propto P(A \mid I, C, Q) \cdot \prod_{j=1}^{n} P(A_j \mid I_j, C_j, Q_j).
\label{eq:few_shot}
\end{equation}
We evaluate zero- through five-shot settings in the ablation; three-shot gave the best accuracy-cost trade-off for Qwen-VL-Chat, while GPT-5-Nano was largely insensitive to shot count once captions were in place. The difference is meaningful: for Qwen-VL-Chat, each additional shot up to three brings consistent classification gains, and the model appears to be extracting visual pattern guidance from the exemplar images rather than just formatting cues from the exemplar answers. For GPT-5-Nano, the few-shot exemplars primarily set the response format; the reasoning content comes from the caption. This asymmetry suggests that the optimal shot count is architecture-dependent, and practitioners working with vision-centric models should budget more exemplars than those working with reasoning-focused models.

\subsection{LLM-as-a-Judge Answer Selection}
\label{subsec:llm_judge}

Once the two candidates are ready, a judge LLM scores each against a reference $A_{\text{ref}}$ across criteria $\Omega = \{\omega_1, \ldots, \omega_m\}$:
\begin{equation}
\text{Score}(A) = \frac{1}{|\Omega|} \sum_{\omega \in \Omega} g_{\omega}(A, A_{\text{ref}}),
\label{eq:judge}
\end{equation}
where each $g_\omega \in [0,1]$. The criteria differ by task type. For recognition, we weight pathogen identification correctness, symptom description precision, and crop identification accuracy. For knowledge-type questions, treatment specificity and practical actionability are incorporated as additional scoring dimensions. Response length is not scored. An earlier version of the rubric included it by default; we removed it after tracing approximately 12\% of misselections to verbosity-driven preferences, consistent with \citet{shi2025judging}. Full scoring rubrics for both task types are detailed in Appendix~\ref{appendix:rubrics}.

Selecting the evaluation criteria required iterative refinement. The initial criterion set was broad but internally redundant---ten dimensions that included both ``factual accuracy'' and ``correctness'' as separate items, which the judge consistently assigned near-identical scores, contributing noise rather than discrimination. Reducing to five or six orthogonal dimensions per task type stabilised the scoring distribution and reduced variance in selections across different judge prompt formulations. The critical structural decision was separating domain-specific criteria---pathogen identification accuracy, crop morphological correctness---from generic writing quality criteria such as completeness and clarity. Without this separation, a response that misidentified the pathogen but was well-structured could outscore a technically accurate but less fluent answer. Since an agronomist's primary concern is whether the identified organism is correct, not whether the prose is polished, the scoring weights are calibrated accordingly.

The judge's output is a quadruple $\langle A^\ast, s^\ast, s^-, R \rangle$, where $R$ is a brief rationale explaining why $A^\ast$ was preferred:
\begin{equation}
\text{Judgment}(A^{(1)}, A^{(2)}) = \langle A^\ast, s^\ast, s^-, R \rangle,
\label{eq:judgment}
\end{equation}
GPT-5 serves as the judge in text-only mode: it receives the structured caption, the diagnostic question, the reference answer, and both candidates, but not the original image. This restriction prevents vision-side confounds---where the judge's own image interpretation might override the caption-grounded reasoning---from influencing the selection. In informal feedback sessions, several agronomists indicated that the rationale $R$ was the component they examined first, treating it analogously to a colleague's written justification before acting on any recommendation.

\begin{equation}
A^\ast = \arg\max_{A \in \mathcal{A}} \text{Score}(A).
\label{eq:selection}
\end{equation}

Agreement between judge selections and human expert choices is quantified using Cohen's Kappa:
\begin{equation}
\kappa = \frac{P_o - P_e}{1 - P_e},
\label{eq:kappa}
\end{equation}
where $P_o$ denotes observed agreement and $P_e$ the agreement expected by chance. The human validation study detailed in Section~\ref{subsec:qualitative} yields $\kappa = 0.88$, exceeding the $> 0.8$ threshold that convention associates with substantial inter-rater agreement. Algorithm~\ref{alg:cpj_framework} presents the complete Agri-CPJ inference procedure.

\begin{algorithm}[t]
\caption{Agri-CPJ Framework: Complete Inference Pipeline}
\label{alg:cpj_framework}
\begin{algorithmic}[1]
\REQUIRE Crop disease image $I$, Diagnostic query $Q$, Quality threshold $\tau$, Max iterations $N_{\max}$
\ENSURE Final diagnosis answer $A^\ast$, Evaluation report $R$
\STATE \textbf{// Stage 1: Generative Explanational Captioning}
\STATE Generate initial caption: $C_0 \leftarrow \mathcal{M}_{\text{VLM}}(I, P_{\text{few}})$
\STATE Initialize iteration counter: $n \leftarrow 0$
\WHILE{$n < N_{\max}$}
    \STATE Compute quality score: $s(C_0) \leftarrow \frac{1}{k} \sum_{i=1}^{k} w_i \cdot d_i(C_0)$
    \IF{$s(C_0) \geq \tau$}
        \STATE \textbf{break}
    \ENDIF
    \STATE Generate critique feedback: $\mathcal{R}(C_0) \leftarrow \text{LLM-Judge}(C_0, \mathcal{D})$
    \STATE Refine caption: $C_0 \leftarrow \mathcal{M}_{\text{VLM}}(I, \mathcal{R}(C_0))$
    \STATE $n \leftarrow n + 1$
\ENDWHILE
\STATE Obtain optimized caption: $C^\ast \leftarrow C_0$
\STATE
\STATE \textbf{// Stage 2: Task-Specific Prompt-Based VQA Generation}
\STATE Generate disease-focused answer:
\STATE \quad $A^{(1)} \leftarrow \mathcal{M}_{\text{VQA}}(I, C^\ast, Q \mid \mathcal{V}_{\text{disease}})$
\STATE Generate crop-focused answer:
\STATE \quad $A^{(2)} \leftarrow \mathcal{M}_{\text{VQA}}(I, C^\ast, Q \mid \mathcal{V}_{\text{crop}})$
\STATE
\STATE \textbf{// Stage 3: LLM-as-a-Judge Answer Selection}
\FOR{$i \in \{1, 2\}$}
    \STATE Evaluate multi-dimensional score:
    \STATE \quad $\text{Score}(A^{(i)}) \leftarrow \frac{1}{|\Omega|} \sum_{\omega \in \Omega} g_{\omega}(A^{(i)}, A_{\text{ref}})$
\ENDFOR
\STATE Select best answer: $A^\ast \leftarrow \arg\max_{A \in \{A^{(1)}, A^{(2)}\}} \text{Score}(A)$
\STATE Generate evaluation rationale:
\STATE \quad $R \leftarrow \text{GenerateRationale}(A^{(1)}, A^{(2)}, A^\ast)$
\STATE \textbf{return} $A^\ast$, $R$
\end{algorithmic}
\end{algorithm}

Algorithm~\ref{alg:cpj_framework} gives the complete Agri-CPJ inference procedure. The iterative caption refinement loop gates entry into dual-answer generation and judge-based selection. In total, a single Agri-CPJ inference call involves three to five LLM API requests depending on how many refinement rounds the caption requires---a non-trivial cost per query, but one that we argue is warranted given the stakes of agricultural decision-making.

The algorithm does not update any model weights, store state between queries, or require a GPU at inference time---the pipeline runs entirely on API calls. From a deployment standpoint, this property matters more than it might appear. A cooperative extension service needs only API credentials and a thin wrapper script; there is no model-serving infrastructure to maintain, no GPU allocation to manage, and no dependency on a fixed model checkpoint. Updating to a better caption generator requires changing a single configuration parameter. The stateless architecture removes a class of operational overhead---version management, hardware provisioning, runtime monitoring---that routinely prevents research systems from reaching production use.

\section{Experiments and Analysis}
\label{sec:experiments}

\subsection{Experimental Setup}

Two benchmarks stress different aspects of the framework. CDDMBench~\citep{liu2024multimodal} is purpose-built for crop disease diagnosis: 3,000 test images covering crop and disease type identification (keyword-matching accuracy) plus a 20-question knowledge QA set spanning 10 disease types scored by GPT-4 on a 1--10 scale normalised to 100 points. We follow the original evaluation protocol without modification. AgMMU-MCQs~\citep{gauba2025agmmu} is deliberately broader---767 four-option multiple-choice questions drawn from 116,231 authentic dialogues between growers and USDA-authorised Cooperative Extension experts, covering disease identification (139 questions), insect/pest identification (144), management instructions (161), species identification (145), and symptom description (178). The format shift from open-ended to multiple-choice was unplanned on our side; we ran the same Agri-CPJ pipeline on both, which made AgMMU a useful proxy for out-of-distribution generalisation.

We chose CDDMBench and AgMMU for complementary reasons. CDDMBench's 3,000-image scale and explicit disease-labelling task gives a clean signal for measuring whether captions help with specific identification, while the knowledge QA component tests whether the framework can support open-ended reasoning rather than just classification. AgMMU's multiple-choice format, broader scope, and origin in real grower--extension expert dialogues provide a very different evaluation surface: the questions were not designed to test any particular model, and the correct answers reflect practical agricultural knowledge rather than benchmark-construction choices. The fact that we did not adapt the pipeline between benchmarks---same prompts, same hyperparameters---makes the AgMMU results a genuine test of how well the approach generalises rather than how well we tuned it.

For CDDMBench, our comparison points are: the original Qwen-VL-Chat (7B) zero-shot result from \citet{liu2024multimodal} (28.40\% crop, 5.00\% disease, 41 QA), which serves as the published anchor; our own independently reproduced Qwen-VL-Chat baseline (28.55\% crop, 5.80\% disease, 41.5 QA), run to verify experimental conditions match the benchmark protocol; and a GPT-5-Nano zero-shot baseline we established ourselves (47.00\% crop, 11.00\% disease, 65 QA). For AgMMU-MCQs, we compare against the full leaderboard reported in \citet{gauba2025agmmu}. Proprietary models: GPT-4o (85.25\%), Gemini 1.5 Pro (80.42\%), Claude 3 Haiku (62.00\%). Open-source models: LLaMA-3.2 (73.32\%), LLaVA-OneVision-8B (72.53\%), LLaVA-1.5-13B (66.73\%), LLaVA-NeXT-8B (66.71\%), Cambrian-8B (65.81\%), LLaVA-1.5-7B (64.16\%), VILA1.5-13B (63.00\%), Qwen-VL-7B (62.34\%), InternVL2-8B (60.17\%).

CDDMBench classification accuracy is determined by keyword-matching: the model's response must contain the correct crop or disease name as it appears in the ground-truth annotation. Knowledge QA is scored by GPT-4 on a 1--10 scale (normalised to 100 points) based on usefulness, relevance, and factual accuracy; we follow the original protocol without modification. For AgMMU, we use exact-match accuracy against the option label (A, B, C, or D). Cohen's Kappa ($\kappa$) is computed for the human validation study to quantify agreement between LLM judge selections and expert choices.

Caption generation used Qwen2.5-VL-72B-Instruct and GPT-5-mini, both processed through the same iterative refinement loop at $\tau = 8/10$. For CDDMBench, the caption prompt instructs the model to describe plant morphology, observable disease symptoms, lesion severity and spatial distribution, and any diagnostic uncertainty---without naming the crop species or disease. The same debiasing constraint and pipeline ran on AgMMU without modification. Qwen-VL-Chat parameters: temperature 0.5, max\_tokens 400, top\_p 0.8, max\_retries 3. GPT-5-Nano parameters: reasoning\_effort=medium, verbosity=low, 30\,s timeout, max\_retries 3. These hyperparameters were fixed on a short validation run and not adjusted per-task or per-dataset.

On AgMMU we evaluate four incremental configurations to isolate each component's contribution: (i) zero-shot without captions, as a direct LVLM baseline; (ii) zero-shot with quality-optimised captions, isolating the captioning effect alone; (iii) three-shot prompting with caption-enhanced exemplars randomly drawn from the training split; (iv) the full Agri-CPJ pipeline with dual-answer generation and judge selection. The judge was GPT-5 in text-only mode at temperature 0, with no image access during the selection step to avoid vision-side confounds. All API calls go through LangChain~\citep{annam2025langchain,topsakal2023creating,dave2025learning} at temperature=0 throughout for reproducibility. The JSON data schema for each pipeline stage, including field definitions and an end-to-end example, is provided in Appendix~\ref{appendix:dataschema}.

\subsection{Main Results on CDDMBench}
\label{subsec:main-results}

Table~\ref{tab:crop-disease-results} summarises performance on CDDMBench's diagnosis and knowledge QA tasks for both VQA models across caption strategies.

With GPT-5-mini captions, GPT-5-Nano reaches 63.38\% crop accuracy, 33.70\% disease accuracy, and 84.5 on knowledge QA---gains of +22.7 pp and +19.5 points over the no-caption baseline. Qwen2.5-VL-72B captions lag behind for this model (53.20\% crop, 32.80\% disease, 76 QA) but still represent a large improvement over the uncaptioned starting point. The performance gap between the two caption generators did not follow the expected direction: Qwen2.5-VL-72B's larger parameter count did not translate into better downstream accuracy for GPT-5-Nano.

The judge stage contributes a consistent terminal increment for both model configurations. Selected answers averaged 4.9/5.0 under the multi-criteria rubric, against 3.6/5.0 for the rejected alternatives---a 1.3-point gap that indicates the judge is discriminating on substantive quality rather than surface features or positional artefacts.

The two models exhibit substantially different response profiles across framework components, suggesting that the gains arise through distinct mechanisms rather than a single shared process. For GPT-5-Nano, the dominant gain tracks caption quality: the +22.7 pp improvement in disease classification corresponds closely with the semantic grounding supplied by the caption, and appending few-shot demonstrations actually reduces crop classification accuracy by 1.40 pp. This pattern is consistent with a reasoning-oriented model for which well-formed captions already supply sufficient domain grounding, and for which in-context demonstrations introduce distributional noise rather than useful signal. Qwen-VL-Chat presents the inverse case. Captions alone shift disease classification by only +6.30 pp, but adding three-shot exemplars stacks a further +24.84 pp on crop classification. That trajectory fits a model that extracts visual pattern alignment cues from demonstrated images rather than performing explicit verbal reasoning over caption content---a meaningfully different mechanism with distinct implications for deployment in settings where curating high-quality exemplars carries a cost.

The choice of caption generator also matters, and the direction of the effect depends on the downstream model. For GPT-5-Nano, GPT-5-mini captions outperform Qwen2.5-VL-72B captions by +10.18 pp in crop classification despite the Qwen2.5-VL-72B model being substantially larger. Inspecting individual captions, GPT-5-mini outputs more precise botanical vocabulary with fewer hedging phrases---the difference between ``possibly consistent with fungal damage'' and ``orange-brown urediospore masses on the abaxial surface near major veins'' is the difference between a cue that focuses reasoning and one that obscures it. For Qwen-VL-Chat, the relationship reverses: Qwen2.5-VL-72B captions produce better downstream results, suggesting the vision-centric model may be responding to stylistic or structural patterns in the caption rather than its semantic content alone.

\begin{table}[!t]
    \centering
    \caption{Performance comparison on CDDMBench dataset. Baseline methods use zero-shot without caption enhancement. Our Agri-CPJ framework progressively adds optimized captions, few-shot prompting, and LLM-as-a-Judge selection. Best results for each model in \textbf{bold}.}
    \label{tab:crop-disease-results}
    \begin{adjustbox}{max width=\textwidth}
    \renewcommand\tabcolsep{6pt}
        \begin{tabular}{llccc}
        \Xhline{0.7pt}
        \textbf{Model} & \textbf{Method} & \textbf{Crop Classification} & \textbf{Disease Classification} & \textbf{Knowledge QA} \\
        \Xhline{0.7pt}
        \multicolumn{5}{c}{\textit{Caption Generator: Qwen2.5-VL-72B-Instruct}} \\
        \hline
        \multirow{5}{*}{Qwen-VL-Chat} & Zero-shot \citep{liu2024multimodal} Baseline & 28.40\% & 5.00\% & 41 \\
        & Zero-shot (Our Baseline) & 28.55\% & 5.80\% & 41.5 \\
        & ~~+ Caption (Optimized) & 29.30\% & 12.10\% & 46.5 \\
        & ~~~~+ Few-shot & 53.39\% & 24.49\% & 50 \\
        & ~~~~~~+ LLM-as-a-Judge & \textbf{54.90\%} & \textbf{25.39\%} & \textbf{51} \\
        \cdashline{1-5}
        \multirow{4}{*}{GPT-5-Nano} & Zero-shot (Our Baseline) & 47.00\% & 11.00\% & 65 \\
        & ~~+ Caption (Optimized) & 51.20\% & 31.40\% & 75.5 \\
        & ~~~~+ Few-shot & 50.10\% & 31.00\% & 74.5 \\
        & ~~~~~~+ LLM-as-a-Judge & \textbf{53.20\%} & \textbf{32.80\%} & \textbf{76} \\
        \Xhline{0.7pt}
        \multicolumn{5}{c}{\textit{Caption Generator: GPT-5-mini}} \\
        \hline
        \multirow{3}{*}{Qwen-VL-Chat} & ~~+ Caption (Optimized) & 28.99\% & 7.17\% & 44 \\
        & ~~~~+ Few-shot & 51.00\% & 14.50\% & 49 \\
        & ~~~~~~+ LLM-as-a-Judge & \textbf{51.60\%} & \textbf{14.80\%} & \textbf{49.5} \\
        \cdashline{1-5}
        \multirow{4}{*}{GPT-5-Nano} & Zero-shot (Our Baseline) & 47.00\% & 11.00\% & 65 \\
        & ~~+ Caption (Optimized) & 60.30\% & 31.60\% & 84 \\
        & ~~~~+ Few-shot & 58.90\% & 29.80\% & 76 \\
        & ~~~~~~+ LLM-as-a-Judge & \textbf{63.38\%} & \textbf{33.70\%} & \textbf{84.5} \\
        \Xhline{0.7pt}
        \end{tabular}
    \end{adjustbox}
\end{table}

\subsection{Caption Impact Analysis}
\label{subsec:caption-impact}

We isolate the contribution of caption enhancement by comparing model performance with and without it across all four model-generator combinations.

Caption enhancement improves all three metrics in every combination we tested. The absolute magnitudes---+22.7 pp on disease classification for GPT-5-Nano, +23.05 pp on crop classification for Qwen-VL-Chat---are large enough to shift practical deployment decisions, and they arise without any parameter update.

\begin{figure}[!t]
    \centering
    \includegraphics[width=0.94\linewidth]{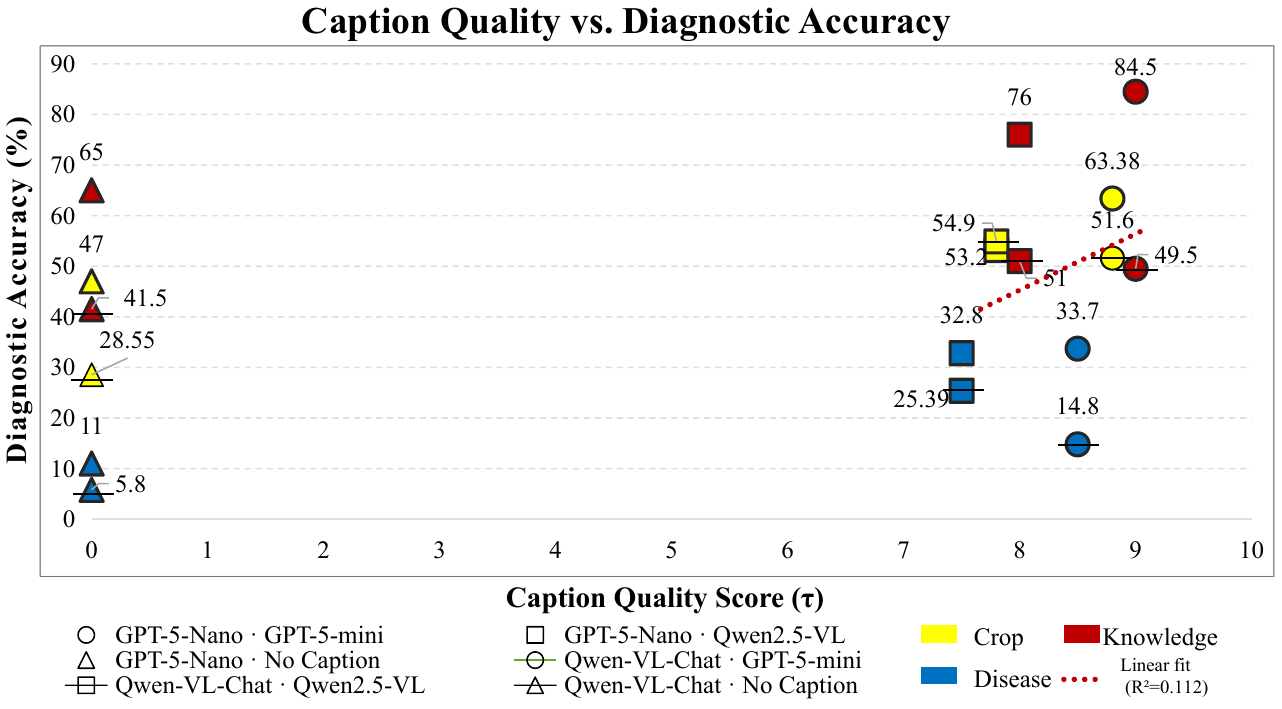}
    \caption{Correlation between caption quality and diagnostic accuracy. The scatter plot shows a positive correlation (R²=0.112) between caption quality scores ($\tau$) and final diagnostic accuracy for our Agri-CPJ framework (Caption + Few-shot + LLM-as-a-Judge). Data points represent GPT-5-Nano (CPJ) and Qwen-VL-Chat (CPJ) across three tasks: Crop Classification (circles), Disease Classification (squares), and Knowledge QA (triangles). Higher-quality optimized captions ($\tau$$\geq$8.5) consistently yield better performance, with an average improvement of $\sim$12 percentage points per quality score increase. The no-caption baseline ($\tau$=0) demonstrates significantly lower accuracy, validating the importance of caption enhancement in our CPJ framework.}
    \label{fig:caption_quality}
\end{figure}

The $R^2 = 0.112$ correlation between caption quality score and downstream accuracy (Fig.~\ref{fig:caption_quality}) is statistically real but modest, indicating that quality is a contributing factor rather than the sole determinant. Decomposing by task reveals an asymmetry: the correlation is strongest in disease classification, whereas crop classification tolerates lower-quality captions more gracefully---consistent with the observation that gross morphological features sufficient for crop identification are often captured even by imprecise descriptions. The most diagnostically valuable caption content is domain-specific terminology for pathogen-characteristic textures: rust pustule colouration, powdery mildew distribution patterns, lesion boundary geometry. A caption reading ``leaves showing some discolouration'' satisfies surface-level quality criteria but contributes little discriminative signal; one reading ``orange-brown urediospore masses concentrated on the abaxial surface near major veins'' provides the anchoring that downstream reasoning requires.

Certain failure cases are worth noting separately. Where most of the leaf surface is already necrotic, captions describe the extent of tissue damage with reasonable accuracy but cannot distinguish between candidate pathogens, because the morphological features that would permit discrimination have been obliterated by disease progression. In such cases, a high-quality caption and a correct diagnosis do not co-occur reliably: the caption may accurately characterise the extent of necrosis while providing no basis for pathogen attribution.

The performance gains are consistent with the architectural account of how captions help. Without captioning, a VLM must extract relevant visual features, translate them into domain terminology, and reason toward a diagnosis---three operations collapsed into one forward pass with no intermediate record. Captioning decouples the first operation: the model produces a text description of what it observes, and that description can be verified against domain knowledge before any diagnostic step proceeds. For GPT-5-Nano, the +22.7 pp disease classification gain from GPT-5-mini captions reflects the difference between a reasoning-capable model receiving well-grounded input versus one that must infer domain vocabulary directly from pixels; the accuracy gap traces to the quality of observation, not to any deficiency in the downstream reasoning process.

\subsection{Ablation Study}
\label{subsec:ablation}

We now quantify how much each Agri-CPJ component contributes by adding them one at a time under the Qwen2.5-VL-72B-Instruct caption setting.

\begin{figure}[!t]
    \centering
    \includegraphics[width=0.95\linewidth]{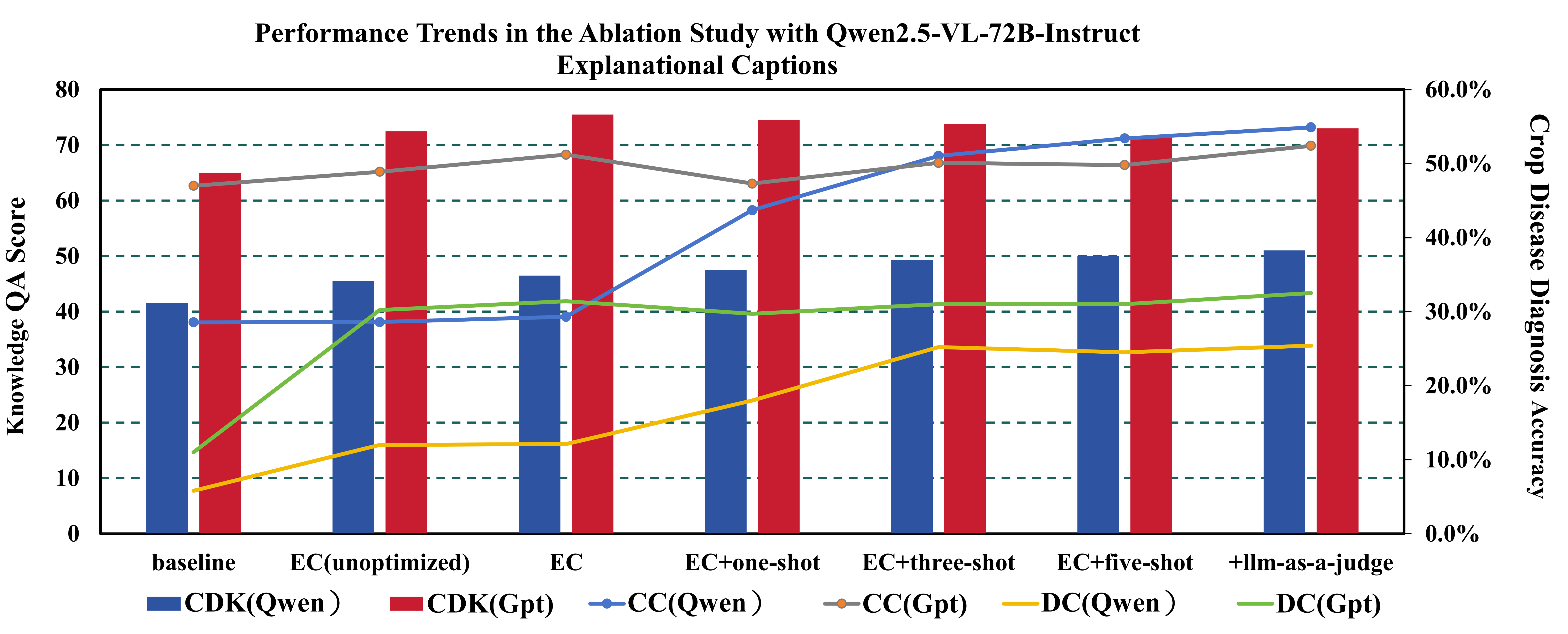}
    \caption{Performance trends in the ablation study using Qwen2.5-VL-72B-Instruct with Explanational Captions (EC). The figure shows progressive improvements from baseline to Agri-CPJ framework across three tasks: Crop Classification (CC), Disease Classification (DC), and Crop Disease Knowledge (CDK). Both Qwen-VL-Chat and GPT-5-Nano demonstrate consistent gains from caption optimization, few-shot prompting, and LLM-as-a-Judge selection, validating the necessity of each component.}
    \label{fig:ablation-study}
\end{figure}

As shown in Fig.~\ref{fig:ablation-study}, progressive gains are visible for both models, though the per-component contribution profile differs in ways that illuminate the underlying mechanisms.

Raw, unrefined captions already produce measurable change. Qwen-VL-Chat gains +6.20 pp on disease classification and +4.0 points on knowledge QA, while crop classification shifts by only +0.05 pp. GPT-5-Nano responds more strongly at the same stage (+19.20 pp disease, +7.5 QA), consistent with reasoning-oriented architectures deriving greater benefit from verbal descriptions---even imperfect ones---than vision-centric models that rely more heavily on direct image encoding.

Iterative quality refinement adds smaller but systematic increments: approximately +0.70/+0.10/+1.0 for Qwen-VL-Chat and +2.30/+1.20/+3.0 for GPT-5-Nano across the three reported metrics. The absolute magnitude is modest, but the function of this step is distinct from merely improving caption fluency. Quality-gating removes near-diagnostic vocabulary that inadvertently encodes ground-truth label information, preserving the integrity of the downstream reasoning process. In our validation set, 35\% of first-pass captions contained at least one such implicit cue; the refinement loop resolved these in 1.8 iterations on average.

Few-shot prompting produces Qwen-VL-Chat's largest single-component improvement: +22.50 pp on crop classification, +19.40 pp on disease classification, and +7.8 points on knowledge QA at three shots. The magnitude and distribution of these gains fit a model that uses demonstrated image--caption--answer triples to learn a visual-to-text alignment function, rather than performing compositional reasoning over caption content---a structurally different process from the mechanism driving GPT-5-Nano's response to captions. GPT-5-Nano is largely unaffected by shot count, consistent with reasoning-focused models benefiting primarily from high-quality inputs rather than diverse demonstrations.

Adding the judge brings the Agri-CPJ framework to 54.90\%/25.39\%/51.0 for Qwen-VL-Chat and 52.40\%/32.50\%/73.0 for GPT-5-Nano. The 2 pp margin from judge-selected answers over arbitrarily chosen candidates appears modest, but the underlying score gap is not: selected answers average 4.9/5.0 versus 3.6/5.0 for rejected ones, confirming that the judge is discriminating on content quality rather than selection order or response length.

\subsection{Qualitative Analysis and Explainability}
\label{subsec:qualitative}

To check whether the judge's automated selections are trustworthy, two PhD plant pathologists independently scored a random 10\% sample ($N=396$) of judged answer pairs from CDDMBench, using the same 5-point rubric. Agreement with LLM selections was 94.2\% (Cohen's $\kappa = 0.88$), average score gap between LLM and expert ratings was 0.23 points, and Pearson $r = 0.91$ ($p < 0.001$). The disagreements we examined tended to cluster on cases where both answers were genuinely near-equal in quality---not cases where the judge was systematically wrong. We are cautious about over-interpreting these numbers: the two experts share a common institutional background, and the sample comes from a single benchmark. Independent replication across additional benchmarks and with expert panels drawn from different institutional backgrounds would strengthen confidence in the $\kappa = 0.88$ figure.

To make concrete what the pipeline produces, consider a tomato leaf presenting early blight symptoms. The refined caption generated by the system reads: ``Compound pinnate leaf exhibiting concentric ring-shaped necrotic lesions (target-spot morphology), dark brown centres surrounded by chlorotic halos; lesions concentrated on older lower leaves, ranging 5--15 mm in diameter, covering approximately 15--20\% of the leaf surface.'' The disease-focused candidate correctly attributes the symptoms to \textit{Alternaria solani} and proposes a fungicide rotation protocol; the crop-focused candidate provides accurate tomato morphology but does not address the pathogen. The judge assigns scores of 4.8 and 4.2 respectively and selects the disease-focused answer, with the following rationale: ``Answer 1 provides precise pathogen identification, detailed symptom characterisation, and an actionable treatment recommendation; Answer 2 accurately describes plant anatomy but offers limited guidance for management decisions.'' A practitioner reviewing that rationale can assess the selection immediately and override it if their field judgement differs---without requiring access to any internal model state.

This traceability---the capacity to locate disagreement at a specific, named component of the output---is the interpretability property the system was designed to deliver. Each Agri-CPJ output comprises four elements: the refined caption with its assigned quality score, both candidate answers with individual criterion scores, and the judge's written selection rationale. These together constitute an audit trail whose interpretation requires no knowledge of the underlying model architecture.

Table~\ref{tab:error_patterns} categorises failure modes across 150 sampled predictions. Early-stage ambiguous presentations account for the largest share (approximately 18\%): when infection has only recently established and the leaf surface remains largely intact, observable symptom features are insufficient for reliable discrimination, and no captioning procedure can recover signal that the image does not contain. Co-infection cases (12\%) present a distinct structural problem---the caption accurately characterises mixed symptom patterns, but the VQA stage is constrained to produce a single-pathogen response, forcing a selection the evidence does not support. Low image quality (8\%) introduces caption inaccuracies through sensor noise and motion blur; pre-processing improvements could address this, though we did not explore them here. Verbosity bias in the judge (5\%) was attenuated but not eliminated by excluding response length from the scoring criteria.

\begin{table}[!t]
    \centering
    \caption{Error pattern analysis with frequency distribution from 150 sampled predictions. The table presents success cases and four primary error categories with their frequencies and characteristics.}
    \label{tab:error_patterns}
    \begin{tabular}{p{0.25\linewidth}p{0.65\linewidth}}
        \Xhline{0.7pt}
        \textbf{Category} & \textbf{Description \& Frequency} \\
        \Xhline{0.7pt}
        Success Case & Bacterial leaf spot: Caption ``necrotic lesions, yellow halos'' $\rightarrow$ correct diagnosis (9/10 QA score) \\
        \hline
        Early-stage Ambiguous & Minimal visual symptoms, insufficient diagnostic evidence ($\sim$18\%) \\
        Co-infection Confusion & Multiple pathogens with overlapping symptoms ($\sim$12\%) \\
        Low Caption Quality & Image artifacts degrade caption accuracy ($\sim$8\%) \\
        Judge Bias & Favors verbose over factually accurate responses ($\sim$5\%) \\
        \Xhline{0.7pt}
    \end{tabular}
\end{table}

Each failure category maps to a specific architectural intervention. The 18\% early-stage failure rate calls for an explicit uncertainty output---a flag indicating that the image does not contain sufficient diagnostic evidence for a confident conclusion---rather than a low-confidence label. Co-infection cases require a multi-label output format evaluated holistically rather than a single-winner selection. The residual verbosity bias requires a stricter factuality weighting in the judge rubric, for which we have not yet identified a stable formulation. These are presented as unresolved engineering problems rather than generic future directions; the modular pipeline structure means they can be addressed as isolated components.

Where the pipeline succeeds, the mechanism is traceable. Explicit captions enable precise pathogen identification by encoding the spatial and morphological observations that diagnostic keys depend on. For bacterial leaf spot, a caption stating ``necrotic lesions with yellow halos following angular patterns along vein boundaries'' provides the information that distinguishes bacterial from fungal infection---including the vein-delimited lesion geometry that is the key discriminating feature. The judge's factuality weighting contributes independently: selected answers achieved 73\% correctness on CDDMBench disease classification versus 51\% for randomly drawn answers from the same dual-answer pool, confirming that the multi-criteria scoring filters content quality rather than preferring surface features such as length or fluency.

\subsection{Cross-Dataset Generalization on AgMMU-MCQs}
\label{subsec:agmmu-generalization}

Table~\ref{tab:agmmu-overall} presents results on AgMMU-MCQs, which tests generalisation to a broader set of agricultural topics in multiple-choice format.

\begin{table}[!t]
    \centering
    \caption{Task-specific performance on AgMMU-MCQs (767 samples) across five agricultural tasks. Baseline models from \citet{gauba2025agmmu}. Our Agri-CPJ progressively adds optimized captions, few-shot prompting, and LLM-as-a-Judge selection, achieving competitive performance with GPT-5-Nano reaching 77.84\% average accuracy and Qwen-VL-Chat achieving 64.54\%.}
    \label{tab:agmmu-overall}
    \begin{adjustbox}{max width=\textwidth}
    \renewcommand{\arraystretch}{1.2}
    \begin{tabular}{llcccccc}
        \Xhline{0.7pt}
        \textbf{Model} & \textbf{Method} & \textbf{Disease/Issue Id.} & \textbf{Insect/Pest Id.} & \textbf{Species Recog.} & \textbf{Management Instr.} & \textbf{Symptom/Visual Descr.} & \textbf{Average} \\
        \Xhline{0.7pt}
        \multicolumn{8}{l}{\textit{Proprietary Models}} \\
        GPT-4o & \multirow{2}{*}{Baseline \citep{gauba2025agmmu}} & 80.51 & 84.46 & 86.23 & 90.42 & 84.64 & 85.25 \\
        Gemini 1.5 Pro & & 76.25 & 79.92 & 81.07 & 88.08 & 76.94 & 80.42 \\
        \hline
        \multicolumn{8}{l}{\textit{SOTA Open-Source Models}} \\
        LLaMA-3.2 & \multirow{3}{*}{Baseline \citep{gauba2025agmmu}} & 65.07 & 69.29 & 75.21 & 78.50 & 78.08 & 73.32 \\
        LLaVA-1.5-13B & & 60.18 & 62.76 & 60.71 & 70.78 & 61.63 & 63.00 \\
        Qwen-VL-7B & & 56.24 & 58.31 & 60.65 & 69.64 & 67.70 & 62.34 \\
        \hline
        \multicolumn{8}{l}{\textit{Ours}} \\
        \multirow{4}{*}{GPT-5-Nano} & Zero-shot (No Caption) & 65.47 & 70.14 & 73.79 & 72.67 & 51.69 & 66.23 \\
        & Zero-shot (Caption, Optimized) & 68.35 & 77.08 & 88.82 & 73.79 & 70.79 & 75.88 \\
        & ~~+ Few-shot & 69.78 & 79.86 & 75.17 & 90.06 & 71.35 & 77.31 \\
        & ~~~~+ LLM-as-a-Judge & \textbf{71.94} & \textbf{77.08} & \textbf{92.55} & \textbf{74.48} & \textbf{72.47} & \textbf{77.84} \\
        \cdashline{1-8}
        \multirow{4}{*}{Qwen-VL-Chat} & Zero-shot (No Caption) & 45.71 & 52.78 & 56.85 & 73.01 & 52.51 & 56.48 \\
        & Zero-shot (Caption, Optimized) & 49.64 & 52.78 & 59.31 & 68.32 & 61.24 & 58.67 \\
        & ~~+ Few-shot & 52.52 & 65.28 & 64.83 & 72.67 & 64.61 & 64.28 \\
        & ~~~~+ LLM-as-a-Judge & \textbf{53.96} & \textbf{66.67} & \textbf{73.29} & \textbf{64.83} & \textbf{62.92} & \textbf{64.54} \\
        \Xhline{0.7pt}
    \end{tabular}
    \end{adjustbox}
\end{table}

GPT-5-Nano with the full Agri-CPJ configuration achieves 77.84\%, positioned between LLaVA-OneVision-8B (72.53\%) and Gemini 1.5 Pro (80.42\%) on the published leaderboard. The 7.41 pp deficit to Gemini is not negligible; it plausibly reflects both the parameter scale difference and the fact that the caption generators used in Agri-CPJ are considerably smaller than those underlying leading proprietary systems. Qwen-VL-Chat reaches 64.54\%, approximately level with LLaVA-1.5-7B (64.16\%)---a reasonable comparison given the two models' similar parameter counts.

Aggregate accuracy, however, obscures substantial variation across the five task categories. Species recognition, where diagnostic features are stable morphological properties, responds most strongly to explicit captioning; leaf shape, venation pattern, and stem texture are precisely the observables that well-constructed captions encode. Disease identification, by contrast, remains the hardest category (71.94\%)---symptom overlap between pathogens and the limited field of view of a single leaf photograph constrain achievable accuracy regardless of caption quality. Management instruction sits between these extremes but is further complicated by the rubric calibration problem: treatment recommendations phrased in the approximate terminology that agronomists routinely use are penalised by a factuality-weighted scoring criterion calibrated on precise textbook formulations.

\begin{figure}[!t]
    \centering
    \includegraphics[width=0.7\linewidth]{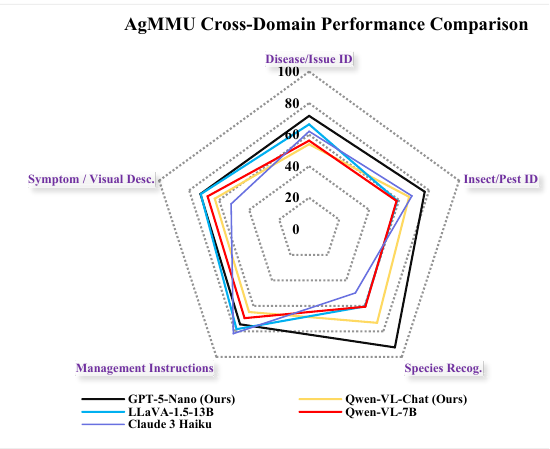}
    \caption{Task-specific performance on AgMMU-MCQs across five agricultural tasks. Our full Agri-CPJ with GPT-5-Nano (CPJ, Ours) and Qwen-VL-Chat (CPJ, Ours) achieves competitive results. GPT-5-Nano reaches highest performance on Species Recognition (92.55\%) and demonstrates strong caption-enhanced reasoning effectiveness across all tasks.}
    \label{fig:agmmu-radar}
\end{figure}

The per-task breakdown (Fig.~\ref{fig:agmmu-radar}) shows one striking outlier. Species recognition climbs from 73.79\% without captions to 92.55\% with the full pipeline---a gain of nearly 19 pp. Plant species identity is almost entirely encoded in morphological features that captions describe explicitly: leaf shape, venation, stem colour, petiole length. Disease identification remains the hardest task at 71.94\%, consistent with CDDMBench. Management instructions follow a non-monotonic trajectory that we cannot fully account for: captions alone give 73.79\%, few-shot pushes it to 90.06\%, but the judge stage drops it back to 74.48\%. The most plausible explanation is that the judge's factuality-weighted rubric penalises treatment recommendations phrased in approximate clinical language---language that is practically sound but less precise than the textbook formulations the rubric was calibrated on.

The CDDMBench and AgMMU evaluations used identical prompts and exemplars---no adjustments were made between the two. Caption enhancement on AgMMU contributed +9.65 pp (66.23\% to 75.88\% for GPT-5-Nano), smaller than the +22.7 pp on CDDMBench disease classification but still substantial. The reduced magnitude is likely partly attributable to format: multiple-choice questions constrain the answer space in a way that partially compensates for weaker visual grounding. The 77.84\% figure was not optimised for AgMMU, which is a relevant point for contextualising it.

\section{Conclusion}

Agri-CPJ was built around one design claim: that forcing visual reasoning into an explicit, quality-checked text representation before any diagnostic question is answered would simultaneously improve accuracy and produce outputs that practitioners can inspect and correct. The experiments reported here support both parts. On CDDMBench, attaching GPT-5-mini captions to GPT-5-Nano yields +22.7 pp in disease classification and +19.5 points on knowledge QA without touching any model weight. On AgMMU-MCQs, the same pipeline achieves 77.84\% with GPT-5-Nano and 64.54\% with Qwen-VL-Chat---results that hold without prompt modifications between the two benchmarks, and that position the framework competitively against open-source models of similar scale. The human validation study ($\kappa = 0.88$) indicates that judge selections track expert preferences closely, though the study draws from a single benchmark and two experts with shared institutional backgrounds; replication with broader expert pools remains necessary.

The failure modes are specific enough to point toward specific fixes. Early-stage images---where infection has only just established and the leaf surface is largely intact---present a detection challenge that no amount of caption quality can resolve; the appropriate system response is an explicit uncertainty signal rather than a low-confidence label. Co-infection cases require multi-label output, not single-winner selection. The management instruction anomaly on AgMMU traces to rubric calibration, not model deficiency. Three to five LLM API calls per query is a real cost at deployment scale, and the pipeline has not been stress-tested against extreme distributional shifts in camera equipment or crop growth stages.

The architectural argument for explicit intermediate representations is not specific to crop disease. Any setting in which visual observations must be rendered in domain vocabulary before reasoning can proceed---medical imaging, ecological monitoring, materials inspection---faces the same semantic gap. We have not tested those domains and make no claim about transfer; but the structural problem Agri-CPJ addresses is not unique to agriculture, and the approach is straightforwardly adaptable.

\section*{Acknowledgments}

This work was supported in part by the Young Scientists Fund of the National Natural Science Foundation of China (NSFC) under Grant 62506084, in part by the Science and Technology Development Fund of Macau SAR (Grant Nos. FDCT/0007/2024/AKP, EF202400185-FST), the UM and UMDF (Grant Nos. MYRG-GRG2024-00165-FST-UMDF, MYRGGRG2025-00236-FST, SHMDF-AI/2026/001).

\bibliographystyle{cas-model2-names}
\bibliography{icme2026references}

\appendix

\section{Prompt Templates}
\label{appendix:prompts}

This appendix documents the key system prompts used across all three Agri-CPJ stages. Core prompts are reproduced verbatim; secondary prompts are summarised with representative excerpts.

\subsection{Stage 1: Caption Generation and Refinement}

\subsubsection{Caption Generation System Prompt}

The following system prompt instructs the VLM (Qwen2.5-VL-72B) to describe plant morphology and disease symptoms without naming any crop species or disease, thereby minimising identification bias in downstream VQA:

\begin{verbatim}
You are an expert agricultural assistant specializing in
describing plant conditions from images.

## Core Task
Describe the visual features of the plant and any disease
symptoms in the image, without identifying the plant or
disease names.

## Key Requirements
1. Focus on describing the plant's morphology, color, and
   overall condition.
2. If disease symptoms are present, describe their appearance:
   color, shape, distribution, size, quantity, and extent.
3. If no disease is visible, state that the plant appears
   healthy.
4. Assess the severity and stage of any symptoms based on
   visual cues.
5. Keep the description concise (90-100 words).
6. If uncertain about features, indicate "unable to describe
   clearly" or "need more images".

## Output Format
{"image_caption": "Description of the plant's visual features
  and any disease symptoms, including morphology, color,
  distribution, size, and condition, without naming the plant
  or disease."}
\end{verbatim}

Three few-shot exemplars accompany the prompt, covering fungal pustules, necrotic lesions, and powdery growth. The fungal pustule exemplar is representative:

\begin{table}[H]
\centering
\caption{Representative few-shot exemplar for caption generation (Stage~1).}
\label{tab:caption-exemplars}
\footnotesize
\renewcommand{\arraystretch}{1.3}
\begin{tabular}{p{0.18\linewidth}p{0.72\linewidth}}
\Xhline{0.7pt}
\textbf{Type} & \textbf{Exemplar Caption} \\
\Xhline{0.7pt}
Fungal pustules &
  The plant has long, slender leaves with numerous orange-brown pustules scattered on the surface. The leaves show premature browning and some collapse. The symptoms suggest a fungal infection that reduces plant vigor, with pustules measuring 1--2\,mm in diameter. The condition appears moderate, affecting approximately 30\% of the leaf area. \\
\Xhline{0.7pt}
\end{tabular}
\end{table}

\subsubsection{Caption Evaluation and Refinement Prompts}

Captions are evaluated by an LLM judge on five criteria: (1)~Accuracy, (2)~Completeness, (3)~Detail, (4)~Relevance, and (5)~Clarity (80--120 words). The judge returns \texttt{rating} (1--10), \texttt{reasoning}, and \texttt{suggestions}. Captions scoring below $\tau = 8.0$ are automatically refined. The refinement prompt instructs the model to address identified issues, add specific symptom details, and maintain professional terminology within 80--120 words, guided by two exemplar captions (\ldots).

\subsection{Stage 2: Dual-Answer VQA Generation}

\subsubsection{Disease Diagnosis System Prompt}

The diagnosis VQA prompt (\texttt{diagnosis\_vqa.py}) instructs the model to return two complementary answers from different diagnostic perspectives. The core rules are:

\begin{verbatim}
...
## Rules
1. Base answers solely on image, caption, and question
2. Prioritize scientific accuracy
3. Never return empty answers
4. BOTH answers must include BOTH plant type and disease type
5. Answer1: Focus on PEST/DISEASE identification
   (symptoms, severity, features)
6. Answer2: Focus on CROP identification
   (type, variety, morphology)
7. Both answers should be scientifically accurate and detailed
\end{verbatim}

The human-turn template is: \texttt{Background(image\_caption): \{image\_caption\}\textbackslash nQuestion: \{question\}}. Five few-shot examples are included; one representative case (apple Alternaria blotch, question: ``Is this crop diseased?'') yields: \textit{Answer~1} identifying Alternaria Blotch with circular brown lesions (2--5\,mm) and yellowish halos, and \textit{Answer~2} identifying the host as \textit{Malus domestica} by ovate leaf shape and serrated margins.

\subsubsection{Knowledge QA System Prompt}

The knowledge QA prompt (\texttt{knowledge\_qa\_vqa.py}) is substantially more detailed, reflecting the open-ended nature of agronomic questions. It defines scope, seven skills, and ten rules governing answer precision:

\begin{verbatim}
You are an agricultural expert specializing in plant disease
diagnosis and management.

Scope: rice blast, tomato leaf mold, wheat leaf rust, apple
gray/sooty/blotch diseases, maize northern leaf blight,
cucurbit powdery mildew, grape leaf blight, tomato yellow
leaf curl virus (TYLCV), pepper bacterial spot, and similar
field/greenhouse pathosystems.

## Skills
1. Extract disease context from background Q&A and question.
2. Diagnose and describe disease mechanisms, signs/symptoms,
   and disease cycles; include differential diagnosis.
3. Translate product names to active ingredients; handle
   dilutions and rates precisely (metric units).
4. Give practical, stage-specific recommendations (seed/
   seedling, vegetative, reproductive), including timings,
   intervals, and number of applications.
5. Integrate IPM: resistant varieties, sanitation, crop
   rotation, canopy management, balanced fertilization
   (N-P-K-Si), irrigation and drainage, environmental
   control (temp/RH/ventilation).
6. Address resistance management: rotate modes of action
   (e.g., FRAC for fungicides, IRAC for insecticides).
7. Ensure scientific accuracy, safety, and applicability.

## Rules (abridged)
1. Base answers on provided background; supplement only with
   widely accepted general agronomic knowledge if needed.
2. If specific data (dose, interval, temperature threshold)
   are provided in the background, use them verbatim.
3. Include active ingredient and formulation when available
   (e.g., 20% tebuconazole EC); provide dilution ratios,
   spray volume, timing, interval, and number of sprays.
4. For cultural practices, include actionable details:
   balanced N-P-K, silicon/potash, drainage, plant spacing,
   pruning for airflow, temperature/RH targets.
5. Rotate fungicides with different FRAC codes; avoid more
   than 2 consecutive applications of the same MoA.
6. Provide TWO answers:
   Answer1: Treatment, prevention, and control (step-by-step
     IPM with specific methods, timings, dosages, intervals).
   Answer2: Disease explanation (symptoms, causes/etiology,
     disease cycle/epidemiology, conducive conditions).
7. Advise PPE, follow local labels, observe pre-harvest
   intervals (PHI) and re-entry intervals (REI).
\end{verbatim}

Five few-shot examples are included. Table~\ref{tab:qa-fewshot} shows one representative case:

\begin{table}[H]
\centering
\caption{Representative few-shot example for knowledge QA (Stage~2).}
\label{tab:qa-fewshot}
\footnotesize
\renewcommand{\arraystretch}{1.3}
\begin{tabular}{p{0.30\linewidth}p{0.60\linewidth}}
\Xhline{0.7pt}
\textbf{Input} & \textbf{Output} \\
\Xhline{0.7pt}
Caption: \textit{Plant: Wheat; Disease: Leaf Rust\ldots} \newline
Question: \textit{What control techniques are applicable to Wheat Leaf Rust?} &
  \textbf{Answer 1 (Treatment):} Control mainly relies on planting resistant varieties, supplemented by chemical treatments. Plant resistant varieties such as Shaanong 7859, Ji 5418. Seed dressing: 0.03--0.04\% triazolone. Foliar spray: 20\% triazolone 1000$\times$ at disease onset, repeat every 10--20 days. Cultural: sow at appropriate time, eliminate volunteers, ensure drainage. \newline
  \textbf{Answer 2 (Explanation):} Wheat Leaf Rust is caused by \textit{Puccinia triticina}. Pathogen overwinters on stubble; spreads via wind-borne urediniospores in spring (optimal 15--22\textdegree C). Multiple infection cycles possible during growing season, leading to premature leaf senescence and yield loss. \\
\Xhline{0.7pt}
\end{tabular}
\end{table}

\subsection{Stage 3: LLM-as-a-Judge Answer Selection}

\subsubsection{Diagnosis Judge System Prompt}

The judge for the disease diagnosis task (\texttt{diagnosis\_judge.py}) evaluates five criteria and returns a structured JSON decision:

\begin{verbatim}
You are an agricultural expert evaluating two answers to a
question about plant disease diagnosis.

## Evaluation Criteria:
1. Accuracy of Plant Identification: Correct identification
   of crop species
2. Accuracy of Disease/Pest Identification: Correct
   identification of disease or pest
3. Symptom Description Accuracy: Precise description of
   disease symptoms
4. Adherence to Required Format: Proper structure with plant
   and disease identification
5. Completeness and Professionalism: Comprehensive and
   scientifically sound response

## Task:
Compare Answer 1 and Answer 2 based on the above criteria
and select the better one.

## Output Format:
{
  "choice": 1 or 2,
  "reason": "Brief explanation for your choice",
  "scores": {
    "answer1": {
      "plant_accuracy": 0-1,
      "disease_accuracy": 0-1,
      "symptom_accuracy": 0-1,
      "format_adherence": 0-1,
      "completeness": 0-1,
      "total": 0-5
    },
    "answer2": { ... }
  }
}
\end{verbatim}

When total scores are within 0.3 points, the tie-breaking rule selects the answer with higher \texttt{plant\_accuracy + disease\_accuracy}. Three few-shot examples are included; one representative case is shown in Table~\ref{tab:diag-judge-fewshot}.

\begin{table}[H]
\centering
\caption{Representative few-shot example for the diagnosis judge (Stage~3).}
\label{tab:diag-judge-fewshot}
\footnotesize
\renewcommand{\arraystretch}{1.3}
\begin{tabular}{p{0.44\linewidth}p{0.46\linewidth}}
\Xhline{0.7pt}
\textbf{Input} & \textbf{Judge Output} \\
\Xhline{0.7pt}
Q: \textit{What disease is affecting this plant?} \newline
Caption: \textit{Apple leaf with Alternaria blotch\ldots} \newline
A1: \textit{Apple leaf with Alternaria blotch. Symptoms include circular brown spots with yellow halos.} \newline
A2: \textit{This leaf might be diseased. It has some spots.} &
  \texttt{\{"choice": 1, "reason": "Answer 1 correctly identifies both plant (apple) and disease (Alternaria blotch) with specific symptoms. Answer 2 is vague.", "scores": \{"answer1": \{"plant\_accuracy": 1.0, "disease\_accuracy": 1.0, "symptom\_accuracy": 0.9, "format\_adherence": 1.0, "completeness": 0.9, "total": 4.8\}, "answer2": \{"total": 1.6\}\}\}} \\
\Xhline{0.7pt}
\end{tabular}
\end{table}

\subsubsection{Knowledge QA Judge System Prompt}

The knowledge QA judge (\texttt{knowledge\_qa\_judge.py}) follows the same output format but evaluates five different criteria suited to open-ended agronomic questions: (1)~Accuracy, (2)~Completeness, (3)~Specificity, (4)~Practicality, and (5)~Scientific Validity. Three few-shot examples follow the same structure as the diagnosis judge (\ldots).

\section{Evaluation Rubrics}
\label{appendix:rubrics}

\subsection{Caption Quality Rubric (10-point scale)}

Captions are evaluated across five dimensions, each contributing to an overall holistic score:

\begin{table}[H]
    \centering
    \caption{Caption quality evaluation criteria (10-point scale).}
    \label{tab:caption-rubric}
    \footnotesize
    \renewcommand{\arraystretch}{1.3}
    \begin{tabular}{p{0.17\linewidth}p{0.47\linewidth}p{0.25\linewidth}}
        \Xhline{0.7pt}
        \textbf{Criterion} & \textbf{Description} & \textbf{Quality Level} \\
        \Xhline{0.7pt}
        Accuracy     & Correct identification of plant features and disease symptoms & 9--10: Excellent \\
        Completeness & Includes plant type, symptoms, severity, and stage           & 7--8: Good \\
        Detail       & Specific measurements, locations, and patterns               & 5--6: Fair \\
        Relevance    & Information useful for downstream diagnosis                  & 2--4: Poor \\
        Clarity      & Professional terminology, 80--120 words                      & 1: Unacceptable \\
        \Xhline{0.7pt}
    \end{tabular}
\end{table}

\subsection{Answer Selection Rubric: Disease Diagnosis (5-point scale)}

Each criterion is scored 0--1 and summed for a total of 0--5:

\begin{table}[H]
    \centering
    \caption{Answer selection criteria for disease diagnosis task.}
    \label{tab:diagnosis-rubric}
    \footnotesize
    \renewcommand{\arraystretch}{1.3}
    \begin{tabular}{p{0.20\linewidth}p{0.44\linewidth}p{0.25\linewidth}}
        \Xhline{0.7pt}
        \textbf{Criterion} & \textbf{Description} & \textbf{Scoring} \\
        \Xhline{0.7pt}
        Plant Accuracy   & Correct crop species identification      & 1.0: Precise; 0.5: Genus; 0.0: Wrong \\
        \hline
        Disease Accuracy & Correct disease/pest identification      & 1.0: Specific; 0.5: Type; 0.0: Wrong \\
        \hline
        Symptom Accuracy & Precise symptom description              & 1.0: Detailed; 0.5: General; 0.0: Inaccurate \\
        \hline
        Format Adherence & Includes both plant AND disease ID       & 1.0: Both; 0.5: One; 0.0: Neither \\
        \hline
        Completeness     & Comprehensive and professional           & 1.0: Complete; 0.5: Partial; 0.0: Minimal \\
        \Xhline{0.7pt}
    \end{tabular}
\end{table}

\subsection{Answer Selection Rubric: Knowledge QA (5-point scale)}

\begin{table}[H]
    \centering
    \caption{Answer selection criteria for knowledge QA task.}
    \label{tab:qa-rubric}
    \footnotesize
    \renewcommand{\arraystretch}{1.3}
    \begin{tabular}{p{0.20\linewidth}p{0.44\linewidth}p{0.25\linewidth}}
        \Xhline{0.7pt}
        \textbf{Criterion} & \textbf{Description} & \textbf{Scoring} \\
        \Xhline{0.7pt}
        Accuracy           & Scientifically correct information              & 1.0: All correct; 0.5: Mostly; 0.0: Errors \\
        \hline
        Completeness       & Covers all relevant aspects                     & 1.0: Comprehensive; 0.5: Partial; 0.0: Minimal \\
        \hline
        Specificity        & Precise details (rates, timings, methods)       & 1.0: Specific; 0.5: General; 0.0: Vague \\
        \hline
        Practicality       & Actionable for farmers                          & 1.0: Practical; 0.5: Somewhat; 0.0: Not useful \\
        \hline
        Sci.\ Validity     & Evidence-based, proper terminology              & 1.0: Rigorous; 0.5: Adequate; 0.0: Questionable \\
        \Xhline{0.7pt}
    \end{tabular}
\end{table}

\section{Pipeline Data Schema}
\label{appendix:dataschema}

Each stage of the Agri-CPJ pipeline consumes and produces JSON records. Table~\ref{tab:data-schema} summarises the fields added at each stage.

\begin{table}[H]
    \centering
    \caption{JSON field evolution across Agri-CPJ pipeline stages.}
    \label{tab:data-schema}
    \footnotesize
    \renewcommand{\arraystretch}{1.3}
    \begin{tabular}{p{0.13\linewidth}p{0.32\linewidth}p{0.44\linewidth}}
        \Xhline{0.7pt}
        \textbf{Stage} & \textbf{Fields Added} & \textbf{Description} \\
        \Xhline{0.7pt}
        Input &
        {\ttfamily question\_id}, {\ttfamily image},\newline {\ttfamily question}, {\ttfamily answer} &
        Unique ID; image path; question text; ground-truth answer \\
        \hline
        Stage 1 &
        {\ttfamily image\_caption}, {\ttfamily rating},\newline {\ttfamily reasoning}, {\ttfamily suggestions},\newline {\ttfamily evaluated}, {\ttfamily optimized} &
        Generated caption; quality score (1--10); evaluation rationale; improvement suggestions; processing flags \\
        \hline
        Stage 2 &
        {\ttfamily generation\_answer1},\newline {\ttfamily generation\_answer2} &
        Disease-focused answer; crop-focused answer \\
        \hline
        Stage 3 &
        {\ttfamily generation\_answer},\newline {\ttfamily selected\_answer},\newline {\ttfamily selected\_score},\newline {\ttfamily unselected\_score},\newline {\ttfamily evaluation\_reason} &
        Final answer; selected candidate (\texttt{answer1}/\texttt{answer2}); scores on 0--5 scale; judge rationale \\
        \Xhline{0.7pt}
    \end{tabular}
\end{table}

A complete end-to-end example record after Stage~3:

\begin{verbatim}
{
  "question_id": "test_0001",
  "image": "leaf.jpg",
  "question": "Is this crop diseased?",
  "image_caption": "Compound pinnate leaf exhibiting dark brown
    circular lesions (3-8 mm) with yellow halos...",
  "rating": 9,
  "generation_answer1": "Yes, bacterial necrotic lesions...",
  "generation_answer2": "Yes, compound pinnate leaf showing...",
  "generation_answer": "Yes, bacterial necrotic lesions...",
  "selected_answer": "answer1",
  "selected_score": 4.7,
  "unselected_score": 3.2,
  "evaluation_reason": "Answer 1 provides specific disease
    identification with detailed symptom description."
}
\end{verbatim}

\end{document}